# Implementation of Parallel Simplified Swarm Optimization in CUDA

Wei-Chang Yeh[1], Zhenyao Liu[1], Shi-Yi Tan[1], Shang-Ke Huang[1]

[1]Integration and Collaboration Laboratory, Department of Industrial Engineering and Engineering Management, National Tsing Hua University, Hsinchu 30013, Taiwan;

Corresponding author: Wei-Chang Yeh(yeh@ieee.org)

**Abstract—** As the acquisition cost of the graphics processing unit (GPU) has decreased, personal computers (PC) can handle optimization problems nowadays. In optimization computing, intelligent swarm algorithms (SIAs) method is suitable for parallelization. However, a GPU-based Simplified Swarm Optimization Algorithm has never been proposed. Accordingly, this paper proposed Parallel Simplified Swarm Optimization (PSSO) based on the CUDA platform considering computational ability and versatility. In PSSO, the theoretical value of time complexity of fitness function is *O (tNm)*. There are *t* iterations and *N* fitness functions, each of which required pair comparisons m times. *pBests* and *gBest* have the resource preemption when updating in previous studies. As the experiment results showed, the time complexity has successfully reduced by an order of magnitude of *N*, and the problem of resource preemption was avoided entirely.

1. INTRODUCTION

Graphics processing unit (GPU) structures had been widely used to accelerate 3D graphics applications and had been considerably applied to various fields such as medical image registration[1], energy efficiency analysis[2], and earth surface modeling [3]. GPU in prospects of processing capability and memory bandwidth are potent, CPU (central processing units) has much less computational abilities. Alternatively, CPU is good at sophisticated task assignments. GPU computes efficiently by using several control units and a vast number of transistors as execution units instead of one sophisticated control units and caches. Fig. 1.1 shows the differences between CPU and GPU in the number of transistors. GPU contains a few simplified calculation cores, and they are grouped into so-called *Single Instruction Multiple Threads* (SIMT) multiprocessors. Each SIMT is run by multiple arithmetic logic units (ALU), which takes in charge of processing parallel data. The parallelism of GPU compensates low computational efficiency drawback of CPU, as Hager et al. described [4].

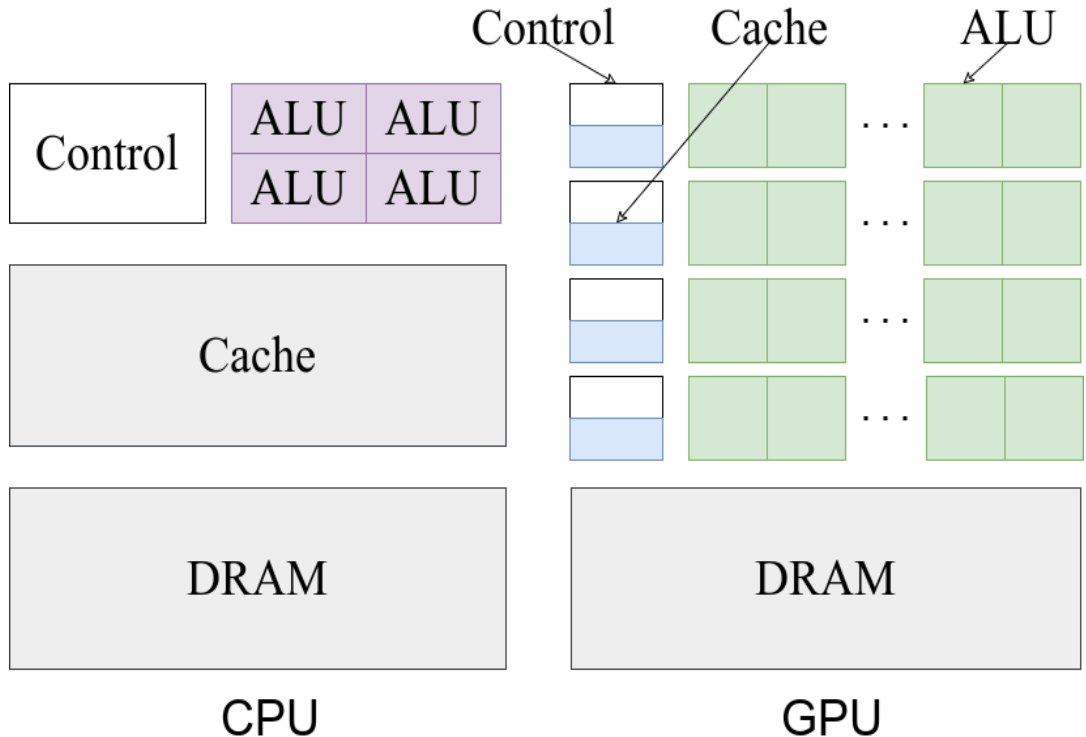

Figure 1.1: The structure of CPU and GPU [5].

GPU of late can support general-purpose data-parallel computation. It also empowered personal computers (PCs) abilities to develop parallel applications with an affordable price

[6] for GPU. OpenCL, Halide, and PGI. are programming languages for GPU. Among parallel programming languages, Nvidia's Computing Unified Device Architecture (CUDA) [5] provides C or C++ language extensions (so-called CUDA-C), which is more intuitive for programming. CUDA-C supports the C and C++ programming language extension, C/C++ programming language are widely used and they can be used to develop a GPU application. Hereafter, CUDA-C and CUDA are used to refer to parallel programming language provided by Nvidia, and these two terms are used interchangeably.

Except for the medical and earth imaging and energy projects, there are more publications used CUDA. Many of them mixed serial control logic and parallel computing data [7]. Using independence-property data is an excellent way to implement parallel computing data. Esma Yildirim et al. [8] defined data transfer throughput as follows: pipelining, parallelism, and concurrency. In aspect of parallelism, Swarm Intelligence (SI) method is inherently data-parallel and data-independent. Additionally, it manipulates data with parallelism instead of concurrency and pipelining. SI has been widely applied to solve real-world problems in engineering and academic domains [9, 10].

Due to distinctiveness of parallelism, the implementation of Swarm Intelligence Algorithms (SIAs) on GPU has obtained great success nowadays. Previous works had shown that SIAs' implementation of GPU rendered CPU-based SIAs' performance significantly [11, 12]. SIAs such as Particle Swarm Optimization (PSO), Genetic Algorithm (GA), Fireworks Algorithm (FWA), their behavior can be boiled down to a decentralized, self-organized swarm (population). Swarm included agents (particles) and acted on the global environment, where agents can associate each other by acting in local environments.

Yeh et al. discussed SSO's parallelism on CPU [13, 14], they had shown that SSO is a data-parallel algorithm that can be implemented on the GPU to compute efficiently. Our research will discuss possible approaches to develop a paralleling SSO on GPU under CUDA platform.

Processors nowadays usually have several cores. The use of parallel programs enables a large number of computational tasks. With the property of parallelism, all data structures that can be precisely split are expected to improve the quality of the solution. This paper proposed Parallel Simplified Swarm Optimization (PSSO), the proposed PSSO which exploits parallelism to improve the quality of solutions compared to CPU-SSO. Moreover, PSSO can be arbitrarily applied to any complex problem in the real world. In PSSO, identical functions are used to solve the split sub-processes. The high degree of independence and similarity of approach between the sub-processes is an excellent strategy for handling them simultaneously.

The remainder of this paper is organized as follows. Section 2 introduces the Simplified Swarm Optimization (SSO) [14], General-Purpose GPU Computing, implementation of the GPU-based SIAs. The proposed PSSO is presented in Section 3 together with the experiments and analysis. Our conclusion is given in Section 4.

## 2. SSO, Gradient Descent, General-Purpose GPU Computing, GPU-based SIAs Implementation

### 2.1. SSO

Yeh proposed SSO in 2009 [14]. As one of SIAs, SSO had shown its efficiency and simplicity to handle real-world problems [15-17]. In 2014, Orthogonal SSO was proposed to solve the parallel redundancy allocation problem (RAP)[18]. However, there had no previous studies proposed GPU-based SSO. SSO got particles at first, which represented the group of potential solutions. It encoded *particles* into proper data structures and stored the structure into a *population* as the second step. These two-steps aimed to initialize the population for further use. SSO started with numbers of random particles. We marked them as *Nsol*. In each particle has *Nvar*, *NSol*, and *Niter*, representing the number of *variables*, *particles*, and *independent replications*, respectively.

$$X_{ij}^{t+1} = \begin{cases} x_{ij}^t & \text{if rand }() \in [0, C_w = c_w) \\ p_{ij}^t & \text{if rand }() \in [C_w, C_p = C_w + c_p) \\ g_j, & \text{if rand }() \in [C_p, C_g = C_p + c_g) \\ x, & \text{if rand }() \in [C_g, 1) \end{cases} \quad (2.1)$$

In each generation, SSO explored the latency dimension using so-called *Heaviside* step function by the given values: *Cw*, *Cp*, and *Cg*. This process was called *SSO Search*, which is the core concept of SSO. Every output value that was transformed from the *stepFunc* function were stored in a *solution*. In order to exploit candidate solutions, SSO in each generation evaluated the best fitness value of a particle was called the personal best value,

*pBests*. With all *pBests* had achieved in all *iters*, the best value in the global space was called *gBest*. Note that the speed plumb dramatically when *Ngen* and *Nsol* were too large. SSO would be unable to find the optima if it was too small for *Nsol* and *Nvar*. The pseudo-code below perfectly draws how SSO worked.

---

**Algorithm 1** SSO algorithm. Good values for the constants are *Nsol* = 50, *Nvar* = 30, *Niter* = 100, $Var_{max}$ = -5.12, $Var_{min}$ = 5.12

---

*sol* = *Nsol* x*Nvar*; *pBests* = *Nsol* x*Nvar*; *gbest* = 0;  ▷Initialize *population*

*Cw* = 0.2; *Cp* = 0.5; *Cg* = 0.8;  ▷Explore Latency dimension

▷Initialize *population*

**while** explorationTime ≤ cpuTimeLimit **do**

  ▷Explore L dimension

  **while** *iter* ≤ *Niter* **do**

    stepFunc (*sol*, *pBests*,*gBest*, randNum($Var_{max}$, $Var_{min}$));

    evaluate(*solF*, *pF*, *gF*);

    **If** *solF* < *pF* **then**

      replace $i^{th}$ *particle* of *pBests* with $i^{th}$ *particle* of *sol*

      **If** *solF* < *pF* **then**

        replace *gBests* with $i^{th}$ *particle* of *sol*

      **end if**

    **end if**

    *Iter* += 1;

  **end while**

**end while**

---

Let $X_{ij}^{t} = X_{i1}^{2}, X_{i2}^{t}, \ldots, X_{iNvar}^{t}$ be the *j*th variables in *i*th particle at iteration *t*. The core idea of SSO is as follows: the current variable is updated by the step function with Eq. 2.1 for exploring latency dimension. To exploiting, the variables were used to search local optima (updating the current value of the *pBests*). The particles were performed the global search (updating global best value, *gBest*), where *solF*, *pF*, and *gF* were representing the fitness value of *sol*, *pBests*, and *gBest*, respectively. The updated current variables maintained the diversity of the population to escape from local optimum.

## 2.2. General-Purpose GPU Computing

The graphics processing unit (GPU) structures have been used to accelerate 3D graphics applications. The characteristics of general-purpose computing on graphics processing units (GPGPU, rarely GPGP) was the capability to make the information bi-directionally (*Heterogeneous Computing*) transferred from GPU to CPU [19]. GPGP was specialized to enhance efficiency with relatively few computational resources on massive size of data. The gigantic-data-level tasks can be parallelized via techno scientifically setups for GPGP. With the infrastructure of GPGP, CPU was responsible for transaction management and the execution of complex logic processing that was inappropriate for data-parallel computing. The GPU comprised mainly of 1-4 memory registers, computational cores, and large cache memory. Thus, GPU was remarkably suitable to address high arithmetic intensity problems and large-scale data-parallel computing by using its processing capability and high bandwidth.

NVIDIA, based on the concept of GPGP, published CUDA in 2007, which supported syntax of C and C++ language. In CUDA programming model, *kernel* served as a bridge for bi-directional communication between CPU and GPU. Each kernel followed SIMT instructions and split it into different *thread group*, which is called *thread block*. When one or more thread blocks were provided for GPU execution, it divided them into *warps*. Since warp executed one SIMT at a time, full efficiency could be achieved when all 32 threads of warp agree to their execution paths. Execution paths can also avail acceleration from memory operations such as fast shared memory or synchronization [5]. CUDA programming model includes three essential functions:

1. Host functions, called and executed only by the host.
2. Kernel functions, it can only be called by the host, but the device executes it. The associated kernel parameters such as *gridDim* and *blockDim* must be specified within the "$<<<...>>>$" when launch it.

3. Device functions can only be called and executed by the device.

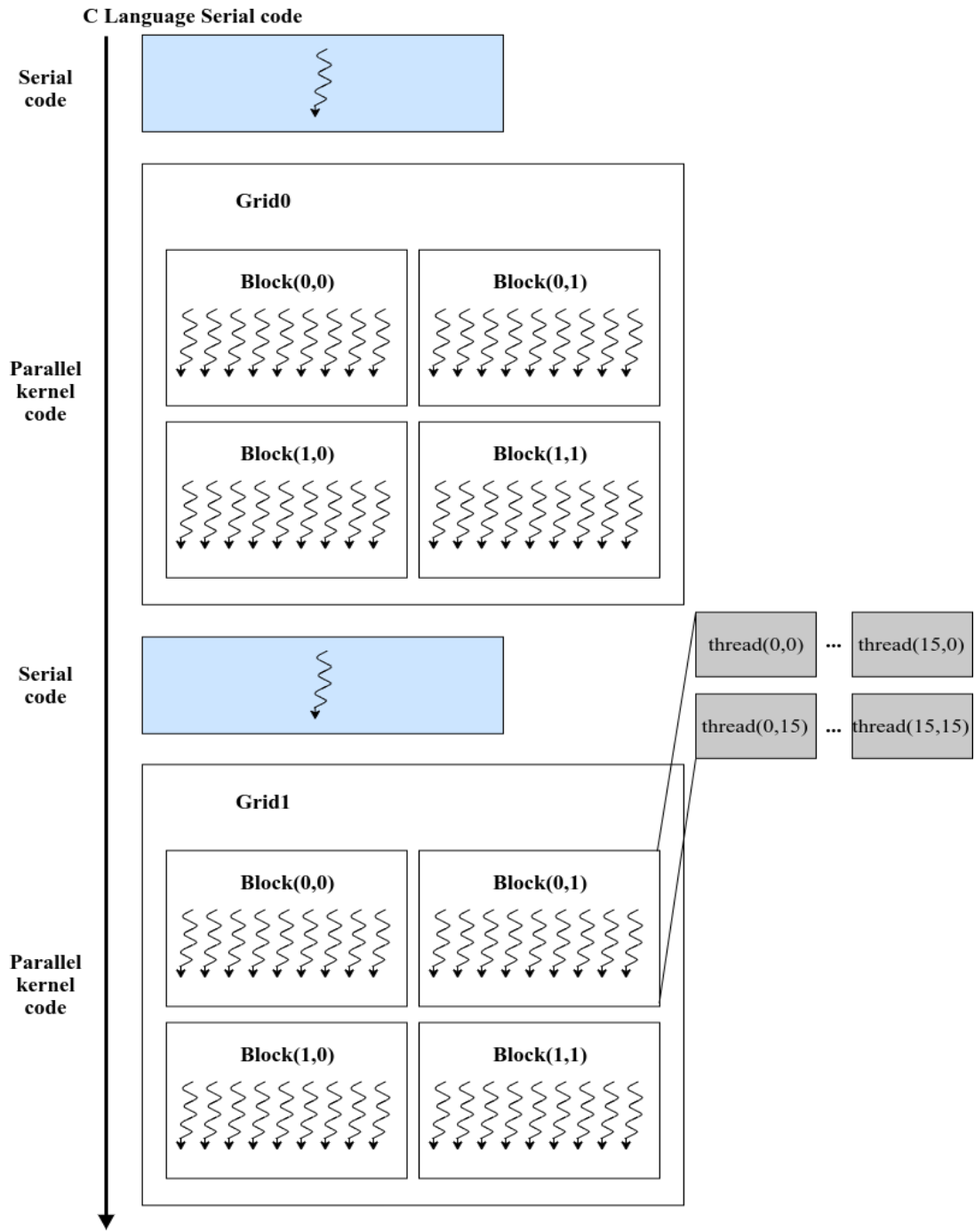

Figure 2.1: CUDA programming model [5].

Note that host functions should be exected as sequential operations on the CPU. Kernel or device functions should be exected as parallel operations on the GPU, but both host and kernel functions must be called on the host [5].

**Memory Model**

CUDA's memory model was closely related to its thread mechanism. There are several types of storage space on the device:

- R/W (Read-write) per-thread registers
- R/W per-thread local memory
- R/W per-block shared memory
- R/W per-grid global memory
- R(Read-only) per-grid constant memory
- R per-grid texture memory

The memory model was illustrated in Fig. 2.2. Threads can only access registers and local memory, shared memory can only be accessed within blocks, and all threads can use global memory in the grid. In this article, we mainly used global memory to achieve.

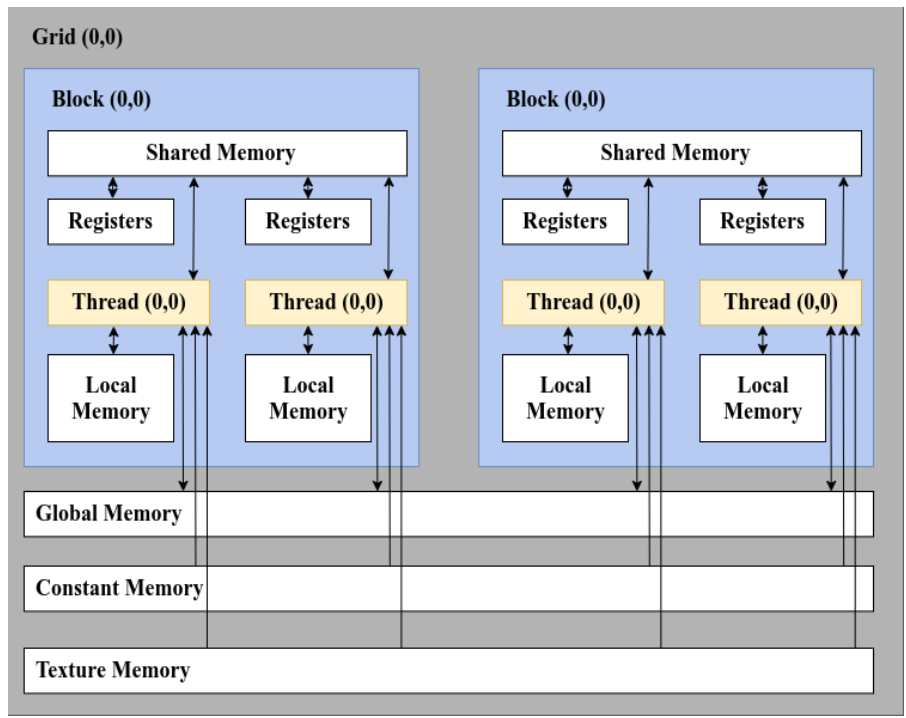

Figure 2.2: Memory Model in CUDA.

### 2.3. GPU-based SIAS Implementation

As a matter of fact, there are several similarities between SSO and other SIAs. In general, the taxonomy of SIAs categorized into four kernel functions [9], Table 2.1 below summarized various GPU-based SIAs:

1. Initialize (I)
2. Evaluate Fitness (E)
3. Communication (C)
4. Update Swarm (U)

In this study, we consider the parallel processing of Kernel Functions to optimize computing resources.

### 2.3.1 Kernel Function (I): Initialize

Kernel Function (I) initialized the population with random numbers and stored them into global memory. Benefit from the intuitive implementation and data access on the global memory, most of the SIAs generated the *population* on CPU [20]. It might have got a vast improvement for computing efficiency if (I) initialize the population on GPU instead of CPU, although the way to arrange the global memory may not be that intuitive [21, 22].

### 2.3.2 Kernel Function (E): Evaluation

Krömer et al. [22] have demonstrated that the most expensive step in SIAs was to evaluate candidate solutions. The most straightforward to deploy Kernel Function (E) is the master-slave paradigm, where the centralized controller dispatched particles in a single population for parallelism. This approach introduced no differences from an algorithmic point of view but did reduce the time-consuming from the computational perspectives. As Tab. 2.1 shows: Zhu et al. [23] implemented a Paralleling Euclidean Particle Swarm Optimization in CUDA. Li et al. [24] proposed a CUDA-based Multichannel Particle Swarm Algorithm. Wong et al. [12] implemented a parallel multi-objective GA. Maitre et al., Tsutsui & Fujimoto [21] ran a sequential SIA and dispatching parallel GA for the particles. According to NVIDIA [5] and Mussi et al. [25], use shared memory in GPU code can guaranty speedup for data transferring. However, most of them did not perform (E) by shared memory.

### 2.3.3 Kernel Function (C): Communication and (U): Update Swarm

Unlike directly distributing the Function (E), Function (C) proffer a more complicated model. It was distinguished by a loosely connected to the population and irregularly exchanging particles. Communicate mechanisms were enabled between swarms according to the law of data access. It means that communication between distributed groups of particles is acceptable. Function (C) and Function (U) have no single pattern to fit all SIAs. We only need to attend the warp divergence, bank conflict in these two Functions.

Table 2.1: Summary of Studies of Taxonomy Analysis for SIAs.

| Reference | SIA | Methodology | Speedup |
| --- | --- | --- | --- |
| Zhou et al. [26] | Stand PSO (SPSO) | (I), (C), (U) on CPU. (E) on GPU without shared memory | ×6 to 8 |
| Zhu et al. [23] | Euclidean PSO (EPSO) | (I), (C), (U) on CPU. (E) on GPU without shared memory | ×1 to 5 |
| Li et al. [24] | Multichannel PSO (MPSO) | (U) on CPU, (I), (E), (C) on GPU without shared memory | ×30 |
| Wong et al. [12] | Multi Objective GA | (I) on CPU, (E), (C), (U) on GPU without shared memory | 10 to 2 |
| Maitre et al. [27] | Coarse Grain Parallelization of GA | (I), (C), (U) on CPU, (E) on GPU only without shared memory | ×60 |
| Qu et al. [28] | Asynchronous and synchronous PSO | (I), (E), (C), (U) on GPU with shared memory | - |
| Tsutsui & Fuji-moto [21] | GA | (I), (E), (C), (U) on GPU with shared memory | ×2 to 12 |
| Krömer et al. [22] | GA and Differential Evolution (DE) | (I), (E), (C), (U) on GPU with shared memory and synchronization | ×3 to 28 for GA, ×19 to 34 for DE |

## 3. PARALLEL SIMPLIFIED SWARM OPTIMIZATION (PSSO)

In this section, PSSO is described, which is a parallel SSO implemented in the CUDA structure. The proposed PSSO was illustrated in Fig.3.1. We carried out the same initialization operation as well as equally manipulated updating and calculating for the best position and the fitness value. Notwithstanding the sequential method for the above, proposed PSSO adopted a parallel approach for Kernel Function (E) with much lower latency in global memory.

### 3.1. Random Number Generation

Optimized performance depended on the quality of random numbers. It is vital to get random numbers in SIAs. Nevertheless, generating a large number of high-quality random numbers can be very time-consuming. To avoid switch time between CPU and GPU, PSSO used the efficient cuRAND library [5] to generate high-quality random numbers on the GPU. In this implementation, the number of random numbers is equal to the product of the population, the variables, and the number of iterations in the swarm.

### 3.2. Thread Organization

In terms of efficiently using powerful and high-speed data access, a straightforward way to implement Function (E) was to use the global memory data access approach [10]. In PSSO, each particle was assigned to one warp (i.e., 32 consecutive threads) or multiple warps within a memory segment, but not all threads were used to perform the Kernel Function (E). For the purpose of using thread-synchronized inherently and threads-communicating within a block, threads were usually in the same warp[29, 30] . However, in PSSO, the number of random numbers was the same as the number of variables multiplying the number of particles. It was usually a tremendous scale far exceeding the maximum memory size that shared memory can tolerate (48KB). In PSSO, the search process value and fitness value of each particle were stored in the global memory but not the shared memory. In order to coalesce global memory access [29], data were usually designed in an interleaved order [30, 31] (See Fig. 3.2). In this implementation, the memory access is aligned and sequential, and hence is coalescing.

Running threads on the same warp need to load data of specific particles from global memory, so particles' data should be stored sequentially. Comparing with the interleaved manner, the organization was more accessible for expanding the scale of the problem. Ultimately, the algorithm was suitable to extend to various forms of the problem since GPU automatically dispatched the number of thread blocks according to memory and computational resources.

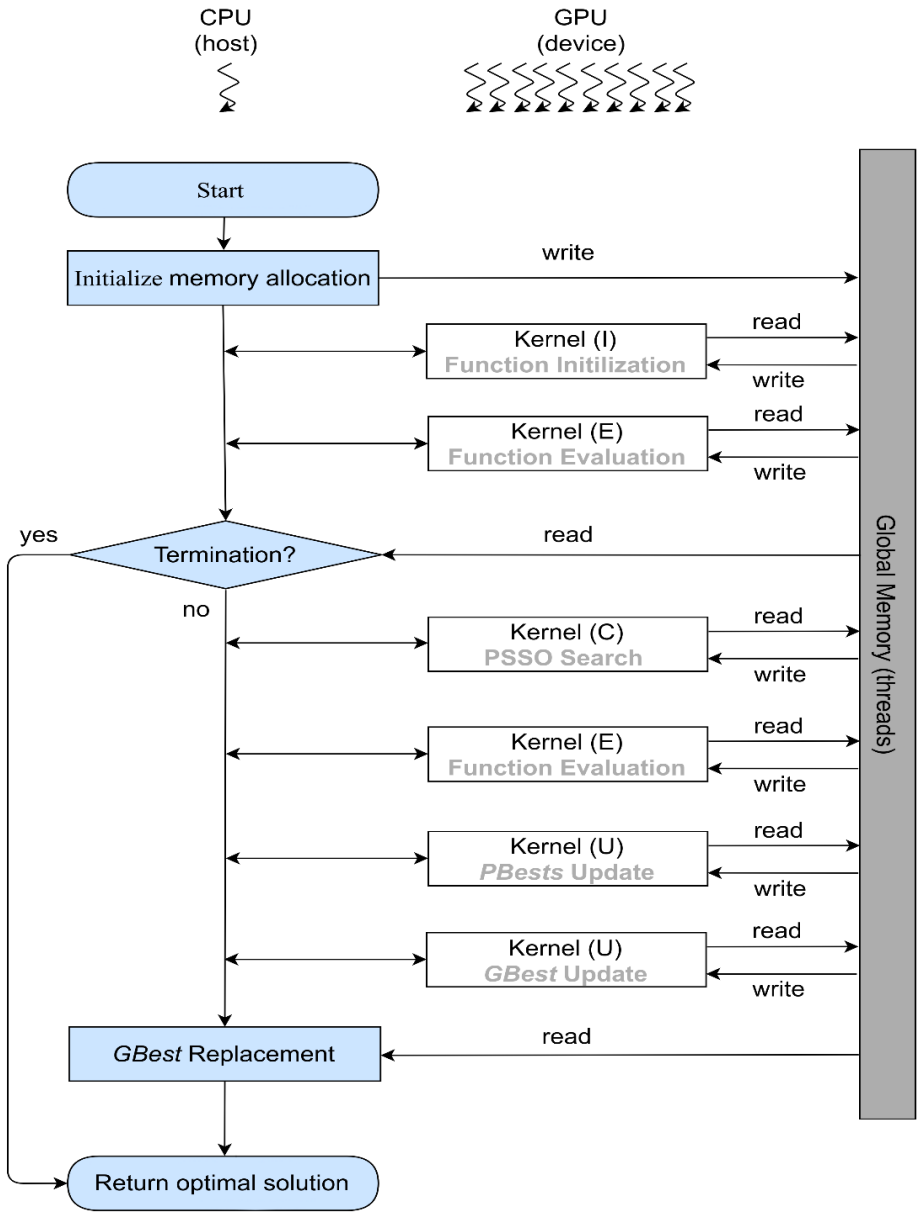

Figure 3.1: Proposed PSSO algorithm.

Table 3.1: Variable Definition and Initialization for PSSO.

| *Describe* | *Variable* |
|---|---|
| Particle's solutions | *Sol* |
| Personal best of particles | *PBests* |
| Global best of particles | *GBest* |
| Parameters of PSSO Search | *Cw*, *Cp*, *Cg* |
| Range for *rand*()>*Cg* | [*varMin*, *varMax*] |
| Number of generations (iterations) | *Niter* |
| Global best value | $fOpt$ |
| Thread blocks | $B$ |
| Single thread block | $T$ |
| Warps | $Warp$ |

### 3.3. PSSO for GPU Computation

Fig. 3.1 depicted PSSO in detail, in order to fully utilize parallelism, the following particular specifications were selected:

1. Transference between CPU and GPU were minimized [29, 30]

2. Coalesced global memory access mechanism is used [30, 31]

3. High-quality random numbers generated by the CURAND library [5]

4. No shared memory was applied to extend the scope of the problem to be more honorable

Kernel Functions were executed multiple times in parallel with threads when called, instead of being executed once, like conventional functions in C language [5]. The kernel functions performed the following operations:

1. Initialize population with Kernel Function (I) on GPU

2. Load data from global memory

3. Execute Kernel Function (E), (C), (U) on GPU

4. Store results in global memory

5. Send data back to GPU

Assuming the problem's fitness function is $f(X)$ in range [L, U]. The dimension of the problem is *Nvar*, and the total number of populations is *Nsol*. We defined problem variables with the one-dimensional array as showing in Tab. 3.1.

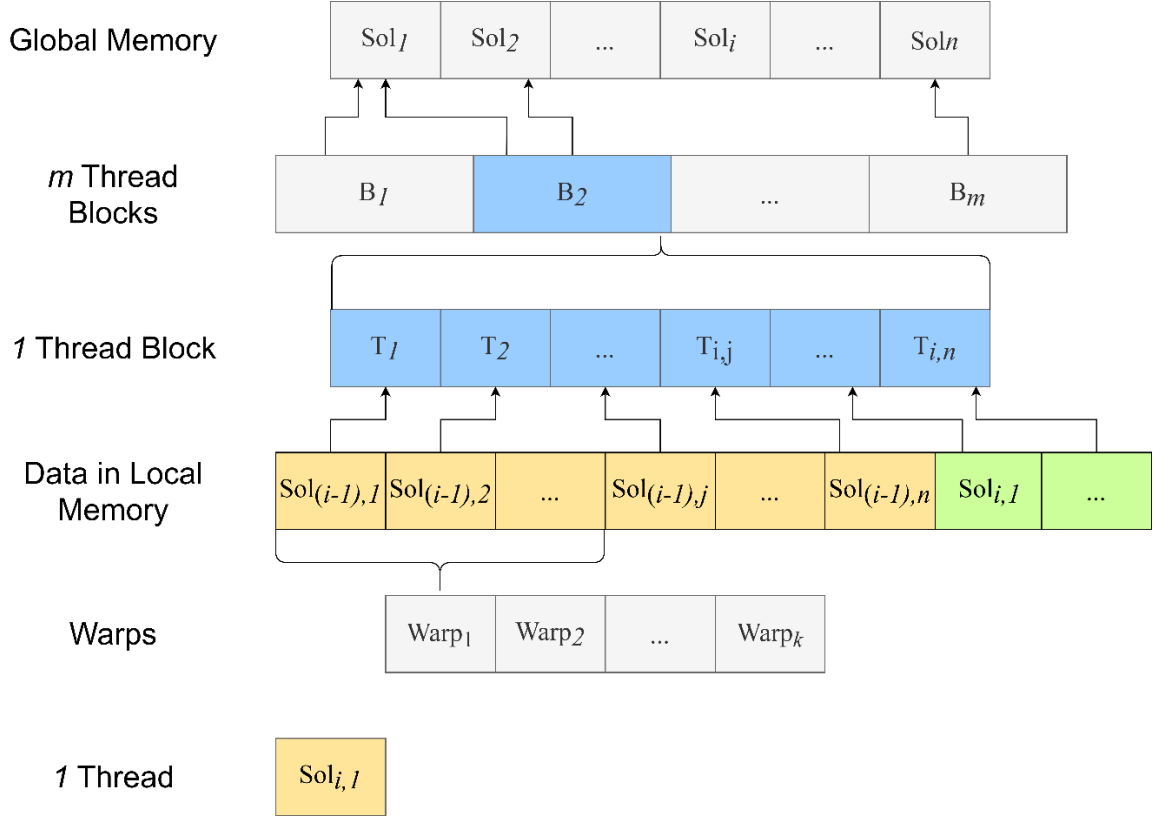

Figure 3.3: Data structure and parallel mode in PSSO.

Fig. 3.3 illustrated that PSSO encapsulated data in a specific way. The solution vectors of all particles were transferred to global memory after Kernel Function (E) finished. Each thread read solution vector of a particle from global memory by thread blocks, and Kernel Function (E) calculated by using specifying threads. Algo.2 showed the algorithm flow of PSSO. *Niter* served as the maximum number of iterations that PSSO to run, which is the terminated condition for the algorithm. One thread in the thread block corresponded to a variable within a solution vector in the population. In the Kernel Function (C), PSSO Search was employed to explore the latent dimension by using Eq. 2.1. Computing fitness function can be invoked only when the thread index equal to the $Sol_{i,0}$.

**Algorithm 2** Algorithmic Flow for PSSO.

*sol = Nsol* x*Nvar*; *pBests = Nsol* x*Nvar*; *Gbest* = 0;     ▷Initialize *population*

set *block size*;     ▷Initialize *block*

__syncThreads();     ▷Wait for *threads* are done

▷Initialize *population*

▷Initialize *block size*

Transfer data from CPU to GPU

// sub-processes in "for" are done in parallel

**for** *iter* = 0 to *Niter* **do**

    Search process for all particles

        ▷__syncThreads()

    Update *PBets* of each particle

    Update *GBest* of each particle

**end for**

Send data back to CPU to calculate *fOpt*

Note that the algorithmic flow is different from Algo.1: In Algo.1, Kernel Function (C), (E), and (U) are executed within a particle. On the contrary, PSSO executed (C) parallel until all particles have done with (C), then do (E) and (U). The difference between CPU and GPU were that we should execute the kernel functions in parallel. Therefore, parallelism must be included for all subprocesses to be optimized. The Kernel Function (E) is the most critical task in the SIAs. To reduce the time complexity, we used the global coalesce memory access method. The way PSSO implemented Kernel Function (U) was shown as in Algo.3 and Algo.4. Multiple threads point to the first element of each particle to parallel perform Kernel Function (U).

**Algorithm 3** Update *Pbests*

__syncThreads();                                                     ▷Wait fo*r threads* are done

**for** each particle *i* **do**

    Map threaads to first element of $i^{th}$ *Sol* and $i^{th}$ *PBests*

Load *Nvar* data from global memory

Calculate data in parallel

**end for**

Paired comparison for $\left(f(Sol_i), f(PBests_i)\right)$

**if** $f(Sol_i) \leq f(PBests_i)$ **then**

    $PBests_i = Sol_i$

**end if**

▷ __syncThreads()

---

**Algorithm 4** Update G*bests*

__syncThreads();                             ▷ Wait for *threads* are done

**for** each particle *i* **do**

    Map threads to first element of $i^{th}$ PBests and GBests

    Load *Nvar* data from global memory

    Calculate data in parallel

**end for**

Paired comparison for $\left(f(PBests_i), f(GBest)\right)$

**if** $\left(f(PBests_i) \leq f(GBest)\right)$ **then**

    $PBests_i = GBests$

**end if**

▷ __syncThreads()

---

**Time complexity**

    Comparing with CPU-SSO, the proposed PSSO has a specific method to encapsulate data, and it has a different algorithmic structure, which took advantage of parallelism to reduce the time complexity in the Kernel Function (U) and PSSO Search process. As Fig 3.4 showed, the computational structure of PSSO differ from CPU-SSO. The CPU only has one thread, and in the CPU-SSO algorithm, the time complexity $O\sim$ is equal to $O(n^3)$. On contrary, proposed PSSO has $O(n)$ and $O(n)$ for CPU and GPU thread(s), respectively. With exponentially decreasing *n* time scale, the run-time of GPU-SSO was expected to be improved significantly, regardless of executing on the architecture of CPU or GPU.

## 3.4. Experiments and Analysis

Large test sets were usually used to compare different algorithms in the field of SIAs, especially when testing involves functional optimization [32]. *No Free Lunch* theorem [33] showed that the performance of any two algorithms would be the same on average if all possible functions had been considered. We have to find the problem type with excellent performance to charac- terize the suitable sort of problem for the algorithm. In this section, the performance of speedup and accuracy will be tested.

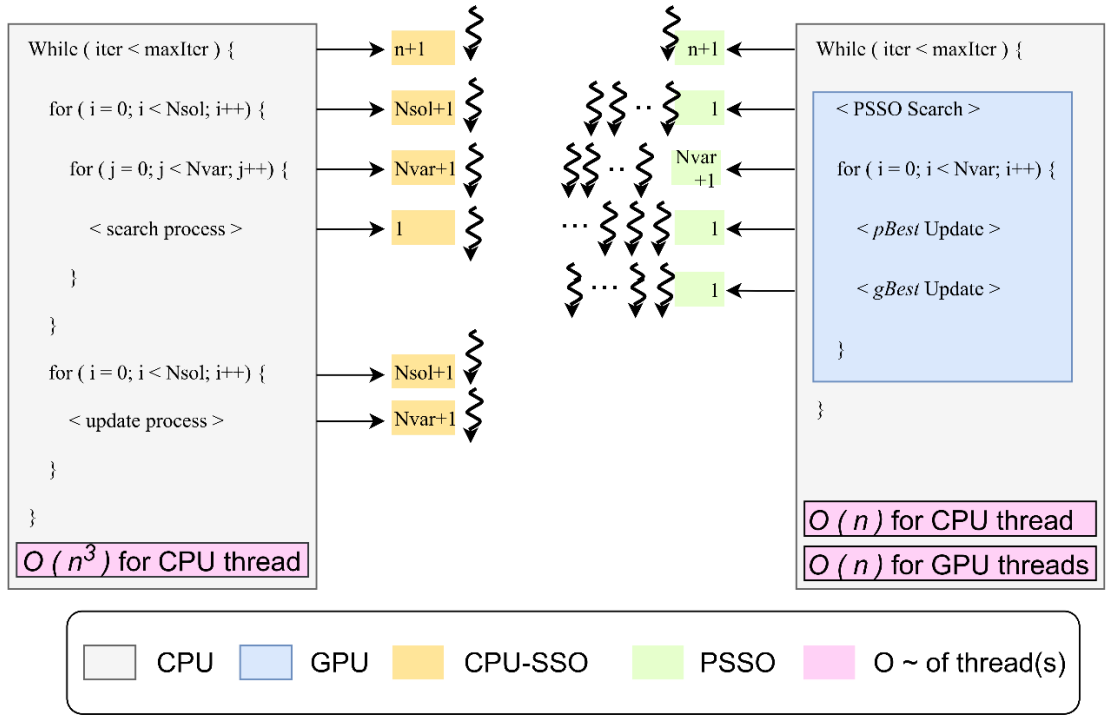

Figure 3.4: Time Complexity for CPU-SSO and PSSO.

### 3.4.1. Benchmark Problems

In order to compare the accuracy and runtime performance of the solution, we implemented two algorithms: CPU-SSO and PSSO. The experiments were conducted on Ubuntu 18.04.1 at the 5.3.0-46-generic kernel version, with 16G DDR3 Memory (1600 MHz) and Intel Core i7- 4770K (3.50GHz × 8 cores, 84W). We used a 180W device card, which is NVIDIA GeForce GTX 1080 with 2560

CUDA cores. The CUDA runtime version is 6.0. Tab. 3.2 showed the definition of each fitness function. The features of each case can be boiled down to: separability, multimodality, and regularity [34].

Table 3.2: Benchmark Functions for PSSO [2].

| ID | Function | Definition | Feasible bounds |
|---|---|---|---|
| $f_1$ | Sphere | $f_1 = \sum_{i=1}^{P} x_i^2$ | $[-5.12, 5.12]^P$ |
| $f_2$ | Hyper-epllipsoid | $f_2 = \sum_{i=1}^{P} i \cdot x_i^2$ | $[-5.12, 5.12]^P$ |
| $f_3$ | Schwefel's | $f_3 = \sum_{i=1}^{P} \left( \sum_{j=1}^{i} \cdot x_j^2 \right)$ | $[-65.536, 65.536]^P$ |
| $f_4$ | Rosenbrock | $f_4 = \sum_{i=1}^{P-1} [100 \cdot (x_{i+1} - x_i^2)^2 + (1 - x_i)^2]$ | $[-2.048, 2.048]^P$ |
| $f_5$ | Rastrigin | $f_5 = 10 \cdot P + \sum_{i=1}^{P-1} [x_i^2 - 10(2\pi x_i)]$ | $[-5.12, 5.12]^P$ |
| $f_6$ | Ackley | $f_6 = -a \cdot \exp\left(-b \cdot \sqrt{\frac{1}{P}\sum_{i=1}^{P} x_i}\right) - \exp\left(\frac{1}{P}\sum_{i=1}^{P} \cos(cx_i)\right)$ | $[-32.768, 32.768]^P$ |
| $f_7$ | Griewank | $f_7 = \frac{1}{4000}\sum_{i=1}^{P} x_i^2 - \prod_{i=1}^{P} \cos\left(\frac{x_i}{\sqrt{i}}\right) + 1$ | $[-600, 600]^P$ |
| $f_8$ | Powell | $f_8 = \sum_{i=1}^{P/4} [(x_{4i-3} + 10x_{4i-2})^2] + [5(x_{4i-1} - x_{4i})^2 + (x_{4i-2} - 2x_{4i-1})^4] [10(x_{4i-3} - x_{4i})^4]$ | $[-4, 5]^P$ |
| $f_9$ | Schwefel | $f_9 = 418.9829 \cdot P - \sum_{i=1}^{P} x_i \cdot \sin\left(\sqrt{|x_i|}\right)$ | $[-5.12, 5.12]^P$ |

In these benchmark functions, $f1, f5, f9$ are separable functions, which means each variable can be optimized in turn. On the other side, inseparable functions are more difficult to optimize because the search direction relies on over one particle. $f1$ to $f3$ are not multimodal functions, while $f4$ to $f9$ are multimodal. The dimension of the search space is also an essential role in the complexity of the problem [35]. In order to set the same difficulty in all problems, we choose a search space of dimension $P = 50$ for all benchmark functions.

### 3.4.2. Design of Experiments

Observing from Tab. 3.3, we know that we need to do a 7-factor experimental design, 128 times of experiments. It is impossible to do such a job in contract computational resources. Thus, the parameters: block size, *Nsol*, *Nvar*, and *maxIter* were arranged as follows: 1024, 100, 50, and 1000, referring to other papers [13, 14, 26, 27, 36].

Table 3.3: Experimental Parameters of PSSO.

| No. | GPU Model  | PSSO                        |
|-----|------------|-----------------------------|
| 1   | Block size | $C_w$, $C_p$, $C_g$         |
| 2   | -          | Population size: *Nsol*     |
| 3   | -          | Particle size: *Nvar*       |
| 4   | -          | Number of iterations: *maxIter* |

The remaining parameters to be tested are the PSSO search parameters: $Cw$, $Cp$, $Cg$, and *Nvar*. Unfortunately, we did not have enough time to find the best one according to the model's running time. Instead, we assigned the following combinations to do experiments (see Tab. 3.4). To set the same difficulty in all problems, first, we have to choose *Nvar*, a search space of dimension $P$ (particle size) for all benchmark functions. Second, using $P$ that gets from the first step to test the performance of PSSO. In this subsection, the experiments are executed by the benchmark function $f_1$.

Table 3.4: Factor for the Parameters of PSSO search.

| No. | $Cw$, $Cp$, $Cg$ |
|-----|------------------|
| 1   | 0.1, 0.3, 0.7    |
| 2   | 0.1, 0.4, 0.8    |
| 3   | 0.2, 0.4, 0.6    |
| 4   | 0.2, 0.5, 0.9    |
| 5   | 0.3, 0.4, 0.5    |
| 6   | 0.3, 0.6, 0.8    |

There are six levels for the PSSO search parameters shown in 3.4, i.e., one of the combi-nations was set to [0,1, 0.3, 0,7] for [$Cw$, $Cp$, $Cg$]. One factor needed to analyze, so One-way ANOVA is chosen. To further analyze, the ANOVA's assumptions listed below have to fulfill.

- **Normality**: The residuals of distribution are normal.

- **Homogeneity of variances**: The variance of data in groups should be the same.

- **Independence of cases**: Each sample and each residual of the samples are independent.

The residuals for $f_1$ are shown in Fig. 3.5. As reported by the *p-value* = 0.0000, which is smaller than 0.05 in the 95% confidence level. We can say that the data do not obey the normal distribution statistically.

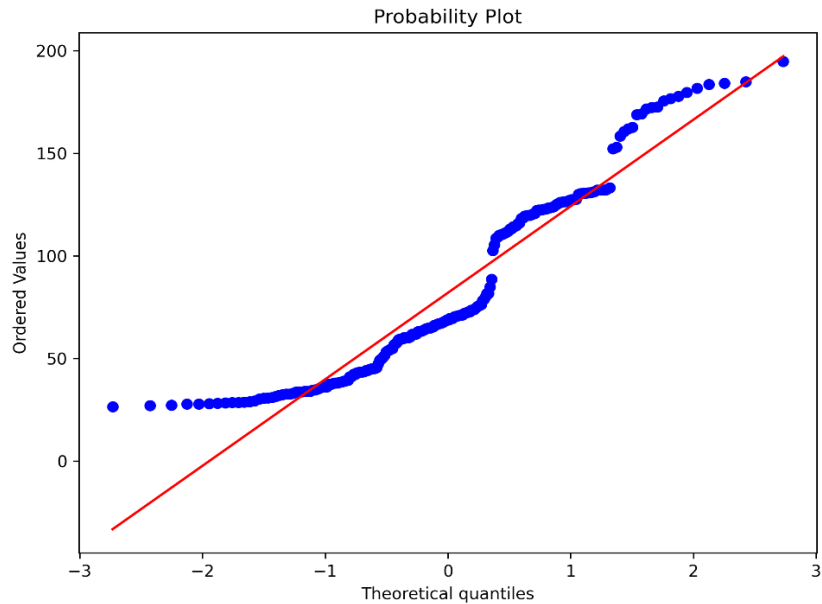

Figure 3.5: Normality Test for Benchmark Function Sphere.

Table 3.5: Homogeneity Test for PSSO.

| *Testing Model* | *Statistic* | *P-value* |
|---|---|---|
| Barlett's Test | 88.6671 | 0.0000 |
| Levene's Test | 34.4380 | 0.0000 |

To test the homogeneity, we adopted Bartlett's test and *Levene's test* to check variance of data. However, to fit input data of both Bartlett's and Levene's testing model, the input data for the normal test were separated into two-part. The size of separated input data was the half size of normality's input data. We used *scipy.stats* library [37] to test the assumption. As Tab. 3.5 explained, both *p-value* were smaller than 0.05 in the 95% confidence level. We have sufficient evidence to refuse the homogeneity assumption. Finally, we adopted the *scipy.stats.ttest_ind* [37] to test independence. By observing Tab. 3.6, the *p-value* is greater than 0.05 in the 95% confidence level, we have enough evidence to reject the data are independent.

Table 3.6: Independence Test for PSSO.

| Testing Model | Statistic | P-value |
|---|---|---|
| T-Test | 15.0809 | 0.0000 |

All assumptions were not reached. The experimental result showed that the data did not obey normality. Also, homogeneity and independence were not fulfilled. Because we did not fulfill the normality, homogeneity, and the independence assumptions, we were going further to analyze the experiments by using *Kruskal-Wallis* H-test. Tab. 3.7 depicted the outperforms performed the best when [$Cw$, $Cp$, $Cg$] was equal to [0.3, 0.6, 0.8]. Besides the points, observing from Tab. A.1 in Appendices A, PSSO achieved better performance when the gap between $Cw$, $Cp$ was set to 0.3. The gap between $Cp$, $Cg$ should not be too large. Otherwise, the outperformance would decrease.

Table 3.7: Kruskal-Wallis H-test for the Parameters of PSSO.

| | $Cw$, | 0.1, | 0.1, | 0.2, | 0.2, | 0.3, | 0.3, |
|---|---|---|---|---|---|---|---|
| | $Cp$, | 0.3, | 0.4, | 0.4, | 0.5, | 0.4, | 0.6, |
| | $Cg$ | 0.7 | 0.8 | 0.6 | 0.9 | 0.5 | 0.8 |
| Method | *Ranking* | 3843.173 | 1968.923 | 4840.817 | 2037.200 | 6270.421 | **1919.306** |
| | *Statistic* | 191.0773 | | | *P-value* | | 2.2989086e-39 |

We implemented CPU-SSO according to section 2.1 and proposed PSSO, as described in section 3. In mimics, we ran $f_1$ to $f_9$ 20 times independently, while 1000 iterations for each run. For CPU-SSO, we performed the same number of function evaluations as PSSO. The two algorithms have been tested on the same criterion with the purpose of making a fair comparison. The experimental parameters were set as follows: P =50, $Cw$=0.3, $Cp$=0.6, $Cg$=0.8. In our experimental environment, the comparison speedup was tested by *Nsol* = 100, 200, 300, and 350 (see section 3.4.3).

### 3.4.3. Precision of Solutions and Speedup.

This subsection showed the trial for CPU-SSO and PSSO in 20 independent runs by testing the benchmark functions (see Tab. 3.2). The average result and corresponding standard deviation were illustrated in Tab. 3.8. We utilized Friedman test [37] to verify differences. As a description in Tab. 3.9, most of the cases have statistical differences for precision of the solutions in PSSO. Additionally, benefit from the algorithmic flow and the data structure that PSSO has (see section 3.3), value of *GBest* was improved significantly. The output data of precision of the solutions for PSSO can be seen from Tab.A.2. In general, as far as the average and the minimum of the performances were concerned, PSSO's performances on multimodal function and unimodal function $f_1$ to $f_9$ worked better than CPU-SSO.

Table 3.8: Precision Comparison for CPU-SSO and PSSO

| Fun. | CPU-SSO | | | PSSO | | |
|---|---|---|---|---|---|---|
| | Avg. | Std. | Min. | Avg. | Std. | Min. |
| $f_1$ | 54.9497 | 7.4781 | 39.0219 | **41.0156** | **5.3095** | **28.5125** |
| $f_2$ | 1152.7869 | 110.1388 | 986.4035 | **820.1844** | **91.6444** | **635.6414** |
| $f_3$ | 192950.2539 | 18823.6598 | 162102.9062 | **127504.9484** | **17093.0233** | **103114.1562** |
| $f_4$ | 1573.8801 | 179.6216 | 1190.2180 | **1103.9103** | **134.5448** | **730.0332** |
| $f_5$ | 269.3232 | **14.4775** | 248.3413 | **220.6183** | 16.2710 | **189.2935** |
| $f_6$ | 16.7117 | **0.2739** | 16.0508 | **15.2896** | 0.3655 | **14.7103** |
| $f_7$ | 199.0340 | 20.2784 | 156.4854 | **145.3612** | **19.4239** | **95.3518** |
| $f_8$ | 1989.3588 | 396.4583 | 1438.9280 | **1181.2840** | **270.4324** | **727.8101** |
| $f_9$ | 20719.6228 | 4.5922 | 20706.0234 | **20708.0471** | **3.6021** | **20702.3574** |

Besides the precision of the solutions, efficiency is a critical factor that also has to be considered. One of the most common measurement methods to compare the test results is *speedup* and *efficiency*. They were illustrated in Eq.3.1 and Eq.3.2. Nevertheless, either speedup or efficiency cannot reflect the exploiting of computational power. Thus, our research adopted performance-criteria: Rectified Efficiency (RE), which was proposed by Tan et al. [9] (Eq.3.3).

$$Speedup = \frac{Time_{CPU}}{Time_{GPU}} \quad (3.1)$$

Table 3.9: Friedman Test for the Precision of the Solutions in PSSO.

| Func. | Statistic | P-Value |
|---|---|---|
| $f_1$ | 19.9200 | **0.0002** |
| $f_2$ | 24.6000 | **0.0000** |
| $f_3$ | 24.6000 | **0.0000** |
| $f_4$ | 21.9600 | **0.0001** |
| $f_5$ | 24.6000 | **0.0000** |
| $f_6$ | 24.9600 | **0.0000** |
| $f_7$ | 21.7200 | **0.0001** |
| $f_8$ | 23.1600 | **0.0000** |
| $f_9$ | 19.5600 | **0.0002** |

$$Ratio = \frac{Power_{GPU}}{Power_{CPU}} \quad (3.2)$$

$$RE = \frac{Speedup}{Ratio} \quad (3.3)$$

Table 3.10: Running Time and Speedup for Benchmark Function Rosenbrock

| Nsol | CPU-SSO | PSSO | RE |
|---|---|---|---|
| 100 | 48.8263 | 0.13875 | 164.2206 |
| 200 | 193.10285 | 0.154 | 585.1602 |
| 300 | 434.8518 | 0.1638 | 1238.8940 |

| | | | | |
|---|---|---|---|---|
| | 350 | 582.71855 | 0.1695 | 1604.3382 |

Speedup experiments were depicted in Tab. 3.10. A series of experiments were carried out to check the speedup of CPU-SSO and PSSO. Among which experiments, the *Nsol* was set to 100, 200, 300, and 350, respectively. The result showed that PSSO accelerates up to ×164.2206 comparing with CPU-SSO when *Nsol* = 100. The speedup's performances were becoming more prominent when size of *Nsol* was increasing. The maximum speedup was ×1604.3382 in the case of *Nsol* = 350.

## 4. CONCLUSION

This dissertation refers to other papers and then proposed a new algorithm: Parallel Simplified Swarm Optimization (PSSO). PSSO was based on SSO algorithm. The core concept of PSSO was to propose a parallel strategy for SSO under CUDA platform. Benefit from different algorithmic flow, CPU-SSO's time complexity as well as precision of the solution were improved significantly statistically. The concept of parallelism was used to enhance the PSSO's performance on GPU. We minimized transference times between CPU and GPU. The logical statements for termination were basically operated on the CPU side. The GPU primarily took charge of arithmetic control to reduce the computational resources efficiently by using the coalesced global memory access mechanism. From the result of experiments, we know that PSSO accelerated up to ×164.2206 comparing with CPU-SSO when the population scale is in 100. The speedup performance was becoming more prominent when size of the population increased. Maximum speedup was ×1604.3382 when population grew up to 350. Furthermore, in aspects of precision, there were statistically different between CPU-SSO and PSSO in all the cases.

## APPENDIX

**Design of Experiments for PSSO**

Table A.1: Input Data of the Normality Test for PSSO.

| | *Cw,* *Cp,* *Cg* | *0.1,* *0.3,* *0.7* | *0.1,* *0.4,* *0.8* | *0.2,* *0.4,* *0.6* | *0.2,* *0.5,* *0.9* | *0.3,* *0.4,* *0.5* | *0,3,* *0.6,* *0.8* |
|---|---|---|---|---|---|---|---|
| CPU | 1 | 71.1325 | 38.0457 | 125.5659 | 28.8096 | 162.6972 | 57.8205 |
| CPU | 2 | 59.5446 | 33.0274 | 124.0904 | 34.0397 | 153.0869 | 39.0219 |
| CPU | 3 | 76.4565 | 27.2271 | 88.7491 | 41.3000 | 181.8000 | 60.3892 |
| CPU | 4 | 60.2476 | 34.0896 | 122.7740 | 29.3240 | 194.8886 | 49.9216 |
| CPU | 5 | 70.9888 | 35.1149 | 132.0811 | 30.6931 | 171.6650 | 53.8014 |
| CPU | 6 | 73.7045 | 32.6563 | 119.4050 | 33.9020 | 158.4920 | 56.8144 |
| CPU | 7 | 81.8001 | 27.9800 | 119.6217 | 30.8272 | 184.1823 | 53.3146 |
| CPU | 8 | 65.0111 | 35.9824 | 127.7038 | 39.3268 | 175.6752 | 47.5388 |
| CPU | 9 | 81.7228 | 28.1868 | 122.6548 | 28.9402 | 168.9867 | 68.9980 |
| CPU | 10 | 75.5534 | 38.5255 | 116.1621 | 30.3440 | 177.7940 | 61.8821 |
| CPU | 11 | 75.9565 | 33.5714 | 119.8743 | 31.0703 | 172.4918 | 62.0330 |
| CPU | 12 | 59.2172 | 31.6079 | 132.1540 | 34.8244 | 169.2028 | 60.3216 |
| CPU | 13 | 73.6397 | 37.5515 | 125.0360 | 28.6832 | 152.2486 | 49.2597 |
| CPU | 14 | 72.8135 | 38.2014 | 130.6304 | 33.8156 | 183.5470 | 63.2274 |
| CPU | 15 | 78.2176 | 32.4620 | 130.4995 | 28.6052 | 176.7337 | 61.9657 |
| CPU | 16 | 72.5103 | 27.8365 | 114.7408 | 32.6657 | 160.7611 | 60.6833 |
| CPU | 17 | 70.4550 | 44.2407 | 114.1584 | 36.2850 | 172.6063 | 51.5327 |
| CPU | 18 | 69.6061 | 27.1518 | 123.0001 | 32.0935 | 161.9723 | 44.8442 |
| CPU | 19 | 66.4468 | 35.7175 | 130.6277 | 26.6378 | 184.8658 | 50.6498 |
| CPU | 20 | 84.9170 | 27.8738 | 130.0180 | 43.6618 | 179.8091 | 44.9746 |
| | *Avg.* | 71.9971 | 33.3525 | 122.4774 | 32.7924 | 172.1753 | **54.9497** |
| GPU | 1 | 122.4114 | 73.9773 | 110.3418 | 64.2969 | 143.5856 | 43.3876 |
| GPU | 2 | 132.4219 | 71.0594 | 123.5448 | 75.3411 | 140.6198 | 45.5630 |
| GPU | 3 | 131.4516 | 70.5970 | 126.5179 | 59.6932 | 138.9908 | 45.0111 |
| GPU | 4 | 131.0374 | 61.7374 | 112.2587 | 72.3286 | 123.9778 | 36.1707 |
| GPU | 5 | 111.6854 | 67.8333 | 118.3873 | 69.5937 | 147.2210 | 43.2663 |
| GPU | 6 | 105.5240 | 63.3601 | 102.6562 | 65.3580 | 134.8146 | 34.5435 |
| GPU | 7 | 123.3983 | 68.9821 | 126.3009 | 78.9082 | 141.8792 | 42.4059 |
| GPU | 8 | 120.1723 | 63.1406 | 120.5428 | 68.5497 | 140.2206 | 42.4614 |
| GPU | 9 | 120.8380 | 65.0306 | 118.5451 | 60.3760 | 150.2198 | 54.1558 |
| GPU | 10 | 127.4057 | 63.8410 | 113.1289 | 64.8807 | 133.3823 | 37.9576 |
| GPU | 11 | 110.9288 | 65.4961 | 108.7186 | 67.2368 | 136.0381 | 28.5125 |
| GPU | 12 | 110.0815 | 63.4492 | 127.1436 | 70.7507 | 147.1328 | 37.5151 |
| GPU | 13 | 109.2608 | 68.1386 | 114.4774 | 74.4213 | 153.2559 | 33.9916 |
| GPU | 14 | 126.5173 | 57.2722 | 126.1160 | 67.3953 | 131.5022 | 38.6262 |
| GPU | 15 | 115.7057 | 66.2593 | 123.7080 | 80.2666 | 152.0076 | 39.5032 |
| GPU | 16 | 133.1897 | 59.6085 | 130.7637 | 66.8367 | 125.7283 | 45.2584 |
| GPU | 17 | 127.4909 | 54.8844 | 122.2111 | 71.2933 | 146.8447 | 43.4856 |
| GPU | 18 | 110.5709 | 54.7824 | 111.2486 | 72.0494 | 146.0819 | 41.1156 |
| GPU | 19 | 119.7704 | 69.6219 | 122.4239 | 64.6893 | 140.8716 | 44.2993 |
| GPU | 20 | 113.3703 | 72.8015 | 132.2346 | 67.0856 | 152.5403 | 43.0814 |
| | *Avg.* | 120.1616 | 65.0936 | 119.5635 | 69.0676 | 141.3457 | **41.0156** |

Table A.2: Output Data of the Precision of the Solutions for PSSO.

| | $f_1$ | $f_2$ | $f_3$ | $f_4$ | $f_5$ | $f_6$ | $f_7$ | $f_8$ | $f_9$ |
|---|---|---|---|---|---|---|---|---|---|
| CPU | 57.82 | 1069.30 | 185060.89 | 1344.71 | 259.23 | 16.95 | 196.72 | 1786.07 | 20718.92 |
| CPU | 39.02 | 1207.01 | 179721.33 | 1579.31 | 260.32 | 16.84 | 234.03 | 1438.93 | 20712.57 |
| CPU | 60.39 | 1010.91 | 231277.19 | 1305.95 | 251.80 | 16.52 | 181.27 | 2504.35 | 20706.02 |
| CPU | 49.92 | 1024.91 | 217473.16 | 1595.23 | 253.40 | 16.88 | 175.41 | 2661.07 | 20722.49 |
| CPU | 53.80 | 993.56 | 234086.36 | 1466.08 | 291.53 | 16.65 | 200.09 | 1911.11 | 20721.33 |
| CPU | 56.81 | 1213.34 | 195778.31 | 1601.14 | 284.36 | 16.58 | 202.84 | 1842.27 | 20721.46 |
| CPU | 53.31 | 1058.56 | 194398.47 | 1479.26 | 263.41 | 16.76 | 203.98 | 2825.84 | 20725.72 |
| CPU | 47.54 | 1361.57 | 190197.06 | 1705.98 | 249.95 | 17.20 | 184.71 | 1768.86 | 20719.32 |
| CPU | 69.00 | 986.40 | 162102.91 | 1647.99 | 269.47 | 16.62 | 213.29 | 1462.96 | 20721.66 |

| | | | | | | | | |
|---|---|---|---|---|---|---|---|---|
| CPU | 61.88 | 1281.48 | 173873.59 | 1536.36 | 286.90 | 16.72 | 189.20 | 1773.87 | 20719.98 |
| CPU | 62.03 | 1256.16 | 184865.73 | 1619.13 | 279.64 | 16.62 | 201.49 | 2083.50 | 20713.47 |
| CPU | 60.32 | 1204.71 | 192596.94 | 1699.70 | 265.15 | 16.40 | 218.22 | 2384.77 | 20722.15 |
| CPU | 49.26 | 1147.99 | 200337.53 | 1679.18 | 284.74 | 17.02 | 197.28 | 1662.26 | 20717.72 |
| CPU | 63.23 | 1041.88 | 212481.92 | 1731.55 | 257.07 | 16.46 | 235.21 | 1544.44 | 20721.78 |
| CPU | 61.97 | 1206.52 | 164635.55 | 1641.58 | 278.79 | 16.41 | 158.47 | 1481.78 | 20717.70 |
| CPU | 60.68 | 1233.61 | 177676.94 | 1190.22 | 285.63 | 16.05 | 200.81 | 2092.47 | 20719.91 |
| CPU | 51.53 | 1261.56 | 200216.28 | 1470.10 | 280.31 | 17.14 | 156.49 | 1922.02 | 20719.85 |
| CPU | 44.84 | 1242.82 | 194000.47 | 1972.64 | 248.34 | 16.89 | 205.72 | 2448.71 | 20727.23 |
| CPU | 50.65 | 1210.58 | 182236.14 | 1385.07 | 251.20 | 16.96 | 215.06 | 2038.71 | 20719.54 |
| CPU | 44.97 | 1042.85 | 185988.31 | 1826.44 | 285.23 | 16.57 | 210.41 | 2153.19 | 20723.66 |
| GPU | 43.39 | 855.83 | 118060.55 | 1063.94 | 189.29 | 15.43 | 156.66 | 1188.65 | 20704.46 |
| GPU | 45.56 | 725.88 | 141413.03 | 1092.53 | 231.59 | 14.71 | 159.39 | 1249.47 | 20707.90 |
| GPU | 45.01 | 967.91 | 131710.67 | 730.03 | 205.58 | 15.26 | 95.35 | 727.81 | 20714.68 |
| GPU | 36.17 | 845.70 | 134990.25 | 1338.93 | 205.13 | 15.05 | 143.00 | 1039.97 | 20709.13 |
| GPU | 43.27 | 939.87 | 129875.73 | 972.40 | 192.81 | 15.21 | 154.22 | 962.32 | 20702.36 |
| GPU | 34.54 | 821.64 | 111364.34 | 1299.02 | 242.38 | 15.21 | 167.35 | 904.35 | 20708.38 |
| GPU | 42.41 | 782.17 | 133603.25 | 1074.08 | 211.55 | 15.45 | 135.07 | 1211.44 | 20710.81 |
| GPU | 42.46 | 739.65 | 108214.88 | 1248.61 | 222.93 | 15.59 | 133.56 | 1332.69 | 20716.58 |
| GPU | 54.16 | 912.20 | 103114.16 | 1077.81 | 227.66 | 15.91 | 170.80 | 1028.79 | 20706.01 |
| GPU | 37.96 | 871.04 | 114409.24 | 1021.06 | 237.07 | 15.30 | 124.51 | 1232.94 | 20705.02 |
| GPU | 28.51 | 860.33 | 130606.30 | 1206.30 | 207.38 | 15.42 | 126.97 | 742.96 | 20710.50 |
| GPU | 37.52 | 916.54 | 137729.39 | 1190.32 | 236.03 | 15.73 | 128.43 | 1276.49 | 20707.72 |
| GPU | 33.99 | 936.44 | 145870.27 | 1209.92 | 220.32 | 15.56 | 156.92 | 925.97 | 20706.60 |
| GPU | 38.63 | 804.80 | 121314.86 | 1177.88 | 225.26 | 14.82 | 147.35 | 1715.38 | 20704.36 |
| GPU | 39.50 | 645.15 | 127713.55 | 1133.51 | 230.44 | 15.76 | 136.74 | 1054.60 | 20707.92 |

| | | | | | | | | |
|---|---|---|---|---|---|---|---|---|
| GPU | 45.26 | 727.07 | 104419.79 | 1066.93 | 254.98 | 14.72 | 159.17 | 1357.28 | 20710.06 |
| GPU | 43.49 | 844.17 | 155562.56 | 914.05 | 228.61 | 14.74 | 180.26 | 1749.65 | 20703.45 |
| GPU | 41.12 | 809.00 | 117463.82 | 1139.54 | 210.75 | 15.54 | 162.61 | 1450.71 | 20711.72 |
| GPU | 44.30 | 635.64 | 111732.80 | 1024.59 | 207.29 | 15.58 | 139.15 | 1406.60 | 20708.95 |
| GPU | 43.08 | 762.64 | 170929.52 | 1096.76 | 225.29 | 14.80 | 129.73 | 1067.60 | 20704.33 |

Table A.3: Output Data of the Speedup Test for PSSO.

| | *Particle Size* | *100* | *200* | *300* | *350* |
|---|---|---|---|---|---|
| CPU | 1 | 49.063 | 191.183 | 437.161 | 564.453 |
| CPU | 2 | 49.073 | 189.712 | 439.999 | 562.614 |
| CPU | 3 | 48.418 | 190.58 | 440.908 | 565.67 |
| CPU | 4 | 47.88 | 192.824 | 437.476 | 563.651 |
| CPU | 5 | 47.758 | 192.861 | 428.533 | 563.799 |
| CPU | 6 | 48.389 | 191.056 | 434.753 | 565.119 |
| CPU | 7 | 49.176 | 188.301 | 434.557 | 571.929 |
| CPU | 8 | 48.248 | 190.205 | 431.854 | 575.904 |
| CPU | 9 | 48.212 | 189.323 | 435.387 | 568.348 |
| CPU | 10 | 50.346 | 189.678 | 432.782 | 582.892 |
| CPU | 11 | 49.061 | 192.337 | 432.366 | 583.594 |
| CPU | 12 | 49.662 | 194.05 | 427.215 | 607.547 |
| CPU | 13 | 49.306 | 195.631 | 429.057 | 601.964 |
| CPU | 14 | 49.547 | 192.663 | 433.167 | 598.993 |
| CPU | 15 | 48.484 | 197.172 | 435.056 | 599.659 |
| CPU | 16 | 48.968 | 196.5 | 432.65 | 598.553 |
| CPU | 17 | 48.827 | 195.68 | 439.168 | 604.617 |
| CPU | 18 | 48.903 | 197.722 | 436.881 | 591.776 |
| CPU | 19 | 47.779 | 196.185 | 439.258 | 594.146 |
| CPU | 20 | 49.426 | 198.394 | 438.808 | 589.143 |
| | *Avg.* | 48.8263 | 193.10285 | 434.8518 | 582.71855 |
| GPU | 1 | 0.15 | 0.166 | 0.18 | 0.19 |
| GPU | 2 | 0.139 | 0.152 | 0.174 | 0.167 |
| GPU | 3 | 0.139 | 0.144 | 0.162 | 0.166 |
| GPU | 4 | 0.138 | 0.144 | 0.169 | 0.174 |
| GPU | 5 | 0.138 | 0.142 | 0.156 | 0.161 |
| GPU | 6 | 0.139 | 0.143 | 0.167 | 0.172 |
| GPU | 7 | 0.141 | 0.148 | 0.157 | 0.16 |
| GPU | 8 | 0.136 | 0.154 | 0.16 | 0.169 |
| GPU | 9 | 0.137 | 0.159 | 0.161 | 0.17 |
| GPU | 10 | 0.136 | 0.146 | 0.166 | 0.165 |
| GPU | 11 | 0.137 | 0.173 | 0.166 | 0.165 |
| GPU | 12 | 0.136 | 0.168 | 0.162 | 0.172 |
| GPU | 13 | 0.137 | 0.153 | 0.163 | 0.169 |
| GPU | 14 | 0.143 | 0.152 | 0.162 | 0.177 |
| GPU | 15 | 0.144 | 0.151 | 0.166 | 0.161 |
| GPU | 16 | 0.135 | 0.16 | 0.165 | 0.177 |
| GPU | 17 | 0.14 | 0.168 | 0.165 | 0.169 |
| GPU | 18 | 0.135 | 0.158 | 0.158 | 0.166 |
| GPU | 19 | 0.134 | 0.15 | 0.158 | 0.17 |
| GPU | 20 | 0.141 | 0.149 | 0.159 | 0.17 |
| | *Avg.* | 0.13875 | 0.154 | 0.1638 | 0.1695 |